\documentclass[conference]{IEEEtran}
\usepackage{cite}
\usepackage{graphicx}
\usepackage{array}
\usepackage{mdwmath}
\usepackage{mdwtab}
\usepackage{eqparbox}
\usepackage[caption=false,font=footnotesize]{subfig}
\usepackage{fixltx2e}
\usepackage{stfloats}
\usepackage{url}
\begin{document}

\title{Complex Networks Measures for Differentiation between Normal and Shuffled Croatian Texts}

\author{\IEEEauthorblockN{Domagoj Margan, Ana Me\v{s}trovi\'c, Sanda Martin\v{c}i\'c-Ip\v{s}i\'c}
\IEEEauthorblockA{Department of Informatics,\\
University of Rijeka,\\
Radmile Matej\v{c}i\'c 2, 51000 Rijeka, Croatia\\
Email: \{dmargan, amestrovic, smarti\}@uniri.hr}}

\maketitle

\begin{abstract}
This paper studies the properties of the Croatian texts via complex networks. We present network properties of normal and shuffled Croatian texts for different shuffling principles: on the sentence level and on the text level. In both experiments we preserved the vocabulary size, word and sentence frequency distributions. Additionally, in the first shuffling approach we preserved the sentence structure of the text and the number of words per sentence. 
Obtained results showed that degree rank distributions exhibit no substantial deviation in shuffled networks, and strength rank distributions are preserved due to the same word frequencies. Therefore, standard approach to study the structure of linguistic co-occurrence networks showed no clear  difference among the topologies of normal and shuffled texts. Finally, we showed that the in- and out- selectivity values from shuffled texts are constantly below selectivity values calculated from normal texts. Our results corroborate that the node selectivity measure can capture structural differences between original and shuffled Croatian texts.
\end{abstract}

\IEEEpeerreviewmaketitle

\section{Introduction}
The complex networks sub-discipline tasked with the analysis of language has been recently associated with the term of linguistic's network analysis. 
The linguistic network can be based on various language constraints: structure, semantics, syntax dependencies, etc. 

In the linguistic co-occurrence complex networks properties are derived from the word order in texts. The open question is how the word order itself is reflected in topological properties of the linguistic network. One approach to address this question is to compare networks constructed from normal texts with the networks from randomized or shuffled texts. Since the majority of linguistic network studies have been performed for English, it is important to test whether the same properties hold for Croatian language as well. So far, there have been only sporadic efforts to model the phenomena of the Croatian language through complex networks \cite{margan2013preliminary} \cite{margan2014shuffled}. 

In this paper we address the problem of Croatian text complexity by constructing the linguistic co-occurrence networks form: a) normal texts, b) sentence-level shuffled texts, and c) text-level shuffled texts. This work extends our previous research \cite{margan2014shuffled} with additional sentence-level shuffling procedure and by introducing a node selectivity as a new complex network measure. Our experiment tests whether  selectivity can differentiate between normal and meaningless texts. 

Section 2 presents an overview of related work on complex network analysis of randomized texts. In Section 3 we define measures for the network structure analysis. In Section 4 we present shuffling procedures and the construction of co-occurrence networks. The network measures are in Section 5. In the final Section, we elaborate the obtained results and make conclusions regarding future work.

\section{Related work}
Some of the early work related to the analysis of random texts dates to 1992, when Li \cite{li1992random} showed that the distribution of word frequencies for randomly generated
texts is very similar to Zipf's law observed in natural languages such as in English. Thus, the feature of being a scale-free network does not depend on the syntactic structure of the language.
Watts and Strogatz \cite{watts1998collective} showed that the network formed by the same amount of nodes and links but only establishing links by choosing pairs of nodes at random has a similar small network distance measures as in the original one. 
Caldeira \textit{et al.} \cite{caldeira2006network} analyzed the role played by the word frequency and sentence length distributions to the undirected co-occurrence network structure based on shuffling. Each sentence is added to the network as a complete subgraph. Shuffling procedures were conducted either on the texts or on the links.
Liu and Hu \cite{liu2008role} discussed whether syntax plays a role in the complexity measures of a linguistic network. They built up two random linguistic networks based on syntax dependencies and compared the complexity of non-syntactic and syntactic language networks.
Krishna \textit{et al.} \cite{krishna2011effect} studied the effect of linguistic constraints on the large scale organization of language. They described the properties of linguistic co-occurrence networks with the randomized words. These properties were compared to those obtained for a network built over the original text. It is observed that the networks from randomized texts also exhibit small-world and scale-free characteristics.
Masucci and Rodgers showed \cite{masucci2006network} \cite{masucci2009differences} that the power law distribution holds when they shuffled the words in the text. Thus, they showed that degree distribution is not the best measure of the self-organizing nature of weighted linguistic networks. Due to the equivalence between frequency and strength of a node, shuffled texts obtain the same degree distribution, but lose all the syntactic structure. They have analyzed the differences between the statistical properties of a real and a shuffled weighted network and showed that the scale-free degree distribution and the scale-free weight distribution are induced by the scale-free strength distribution. They proposed a measure, the node selectivity, that can distinguish a real network from a shuffled network. Selectivity is defined as the average weight distribution on the links of the single node.

Preliminary results of our research on Croatian co-occurrence networks presented in \cite{margan2013preliminary} point out that the increase of the co-occurrence window size is followed by a decrease in diameter, average path shortening and, expectedly, the
condensing of the average clustering coefficient. The stopwords removal causes the same effect. When comparing Croatian literature networks to networks from other languages such as English and Italian \cite{ban2013initial} some expected universalities such as small-world properties are shown, but there are still some differences. The Croatian language exhibits a higher path length than English and Italian language which can be caused by the mostly free word order nature of Croatian.

Initial attempt \cite{margan2014shuffled} to analyse network properties of normal and shuffled Croatian texts show that the text shuffling causes the decrease of the network diameter, due to the establishment of previously non-existing links. Furthermore, the results indicate a slight difference in the average clustering coefficient which is higher for the networks based on the shuffled text. 
Also, obtained results showed that node degree distributions are preserved in text-level shuffled networks, due to the same word frequencies (e.g. Zipf's law).

\section{The network structure analysis}
In the network, $N$ is the number of nodes and  $K$ is the number of links. In weighted networks every link
connecting two nodes has an associated weight $w \in {R}_{0}^{+}$. The co-occurrence window $m_{n}$ of
size $n$ is defined as $n$ subsequent words from a text. The number of network components is denoted
by $\omega$. 

For every two connected nodes $i$ and $j$ the number of links lying on the shortest path between them is
denoted as $d_{ij}$, therefore the average distance of a node $i$ from all other nodes is:
\begin{equation}
d_{i} = \frac{\sum_{j}d_{ij}}{N}.
\end{equation} 

And the average path length between every two nodes $i,j$ is:
\begin{equation}
L = \sum_{i,j} \frac{d_{ij}}{N(N-1)}.
\end{equation}

The maximum distance results in the network diameter:
\begin{equation}
D = max_{i} d_{i}.
\end{equation}

For weighted networks the clustering coefficient of a node $i$ is defined as the geometric average of
the subgraph link weights:
\begin{equation}
c_{i} = \frac{1}{k_{i}(k_{i}-1)} \sum_{ij} (\hat{w}_{ij} \hat{w}_{ik} \hat{w}_{jk})^{1/3},
\end{equation}
where $k_{i}$ is the degree of the node $i$, and the link weights $\hat{w}_{ij}$ are normalized by the maximum weight in the network
$\hat{w}_{ij} = w_{ij}/\max(w)$. The value of $c_{i}$ is assigned to 0 if $k_{i} < 2$. 

The average clustering of a network is defined as the average value of the clustering coefficients of
all nodes in a network:
\begin{equation}
C = \frac{1}{N}\sum_{i} c_{i}.
\end{equation} 

If $\omega > 1$, $C$ is computed for the largest network component.

The out-degree and in-degree $k_{i}^{out/in}$ of node $i$ is defined as the number of its out and in nearest neighbors.

The out-strength and the in-strength $s_{i}^{out/in}$ of the node $i$ is defined as the number of its outgoing and incoming links, that is: 
\begin{equation}
s_{i}^{out/in} =  \sum_{j}w_{ij/ji}.
\end{equation} 

We then define for the node $i$ the out and in selectivity as 
\begin{equation}
e_{i}^{out/in} = \frac{s_{i}^{out/in}}{k_{i}^{out/in}}.
\end{equation}

\section{Methodology}
\subsection{Data}

For the construction and analysis of co-occurrence networks, we used the corpora of 10 books written in or translated into the Croatian language. We divided the corpora into two parts: the first - C1 includes one book, and the second - C2 includes all books. The C1 contains: 191941 words (27453 unique) in 17045 sentences; and C2: 888293 words (91420 unique) in 57179 sentences.

\subsection{The shuffling procedure}
Commonly, the shuffling procedure randomizes the words in the text, transforming the text into the meaningless form. We shuffled the C1 and C2 corpora in two ways: a) shuffling on the sentence level, and b) shuffling on the text level. 
In both ways we preserved the vocabulary size, the word and sentence frequency distributions. 

In the sentence-level shuffling we preserved the sentence length (the number of words per sentence) and sentence order. Versions of C1 and C2 shuffled on the sentence level are noted as C1' and C2'. In the text-level shuffling, the original text is randomized by shuffling the words and punctuation marks over the whole text. This approach preserves the number of sentences but changes the number of words per sentence. Versions of C1 and C2 shuffled on the text level are noted as C1* and C2*. Fig. 1 shows the histograms of sentence length frequencies for C2 and C2* corpora.

\begin{figure}[!t]
\centering
\includegraphics[width=3.5in]{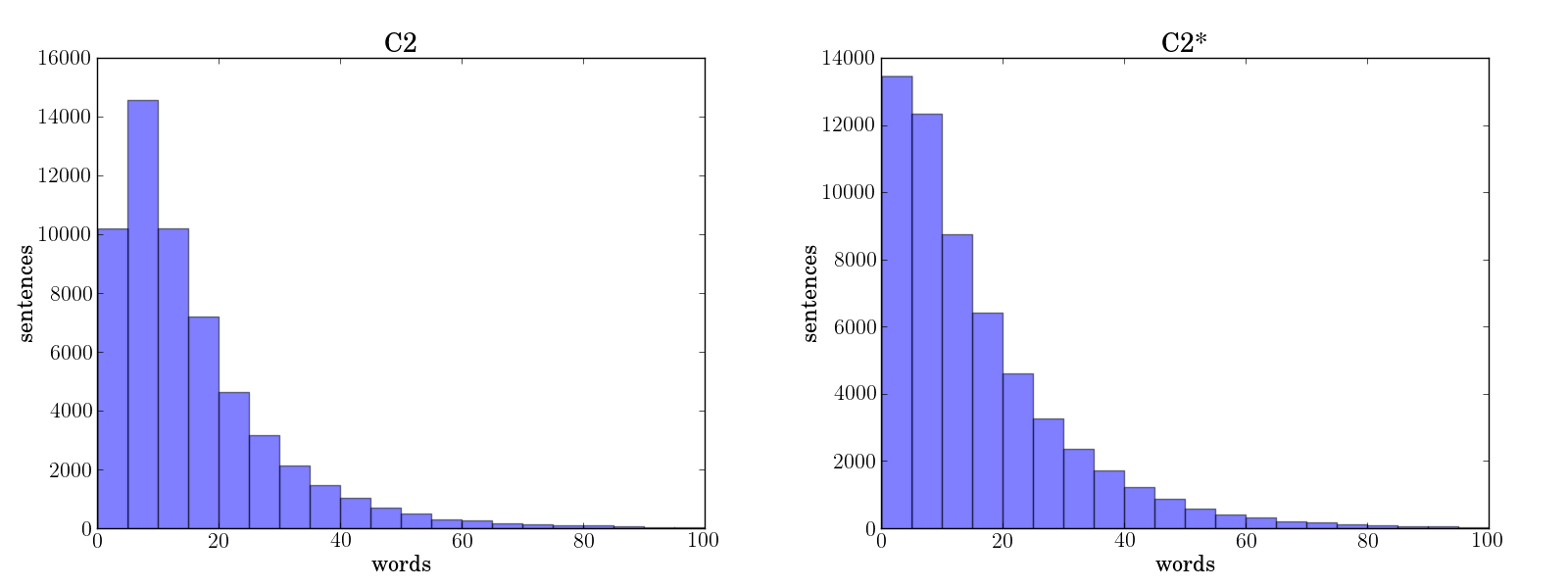}
\caption{Histograms of sentence length frequencies for C2 and C2*}
\label{fig_sim}
\end{figure}

Our shuffling procedures destroy the sentence and text organization in a way that the standard word-order and syntax of the text is eradicated. As expected, the typical word collocations and phrases of Croatian language are completely lost, as well as the forms of the morphological structures and local structures of words' neighborhood. 
 
\subsection{The construction of co-occurrence networks}
Text can be represented as a complex network of linked words: each individual word is a node and interactions amongst words are links. We constructed six different co-occurrence networks (C1, C1', C1*, C2, C2', C2*) all weighted and directed. Words are nodes linked if they are co-occurring as neighbors to each other in a sentence. The weight of the link is proportional to the overall co-occurrence frequencies of the corresponding word pairs within a corpus. 

Network construction and analysis was implemented with the Python programming language using the
NetworkX software package developed for the creation, manipulation, and study of the structure,
dynamics, and functions of complex networks \cite{hagberg2008exploring}. 

\begin{figure}[!t]
\centering
\includegraphics[width=3.5in]{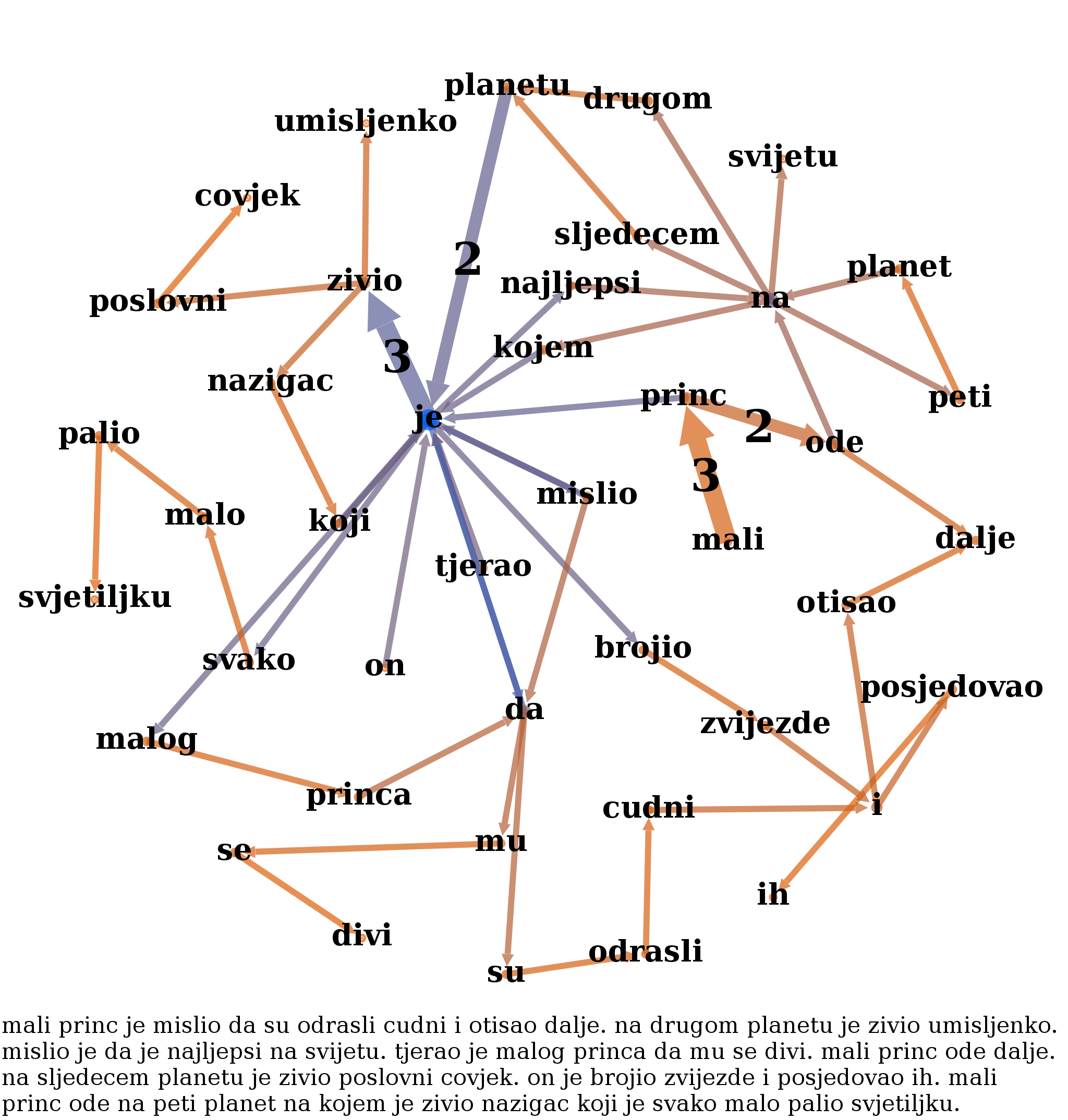}
\caption{Example network constructed from original text}
\label{fig_sim}
\end{figure}

Figures 2, 3 and 4 show simple examples of networks constructed from original and shuffled text samples. Networks are directed and weighted; the numbers on the networks' edges represent weights larger than 1.  At the bottom of each figure is the text from which the visualized network was constructed. The original sample text (Fig. 2) was shuffled on sentence (Fig. 3) and text (Fig. 4) level in the same way as the corpora C1 and C2. 

\begin{figure}[!t]
\centering
\includegraphics[width=3.5in]{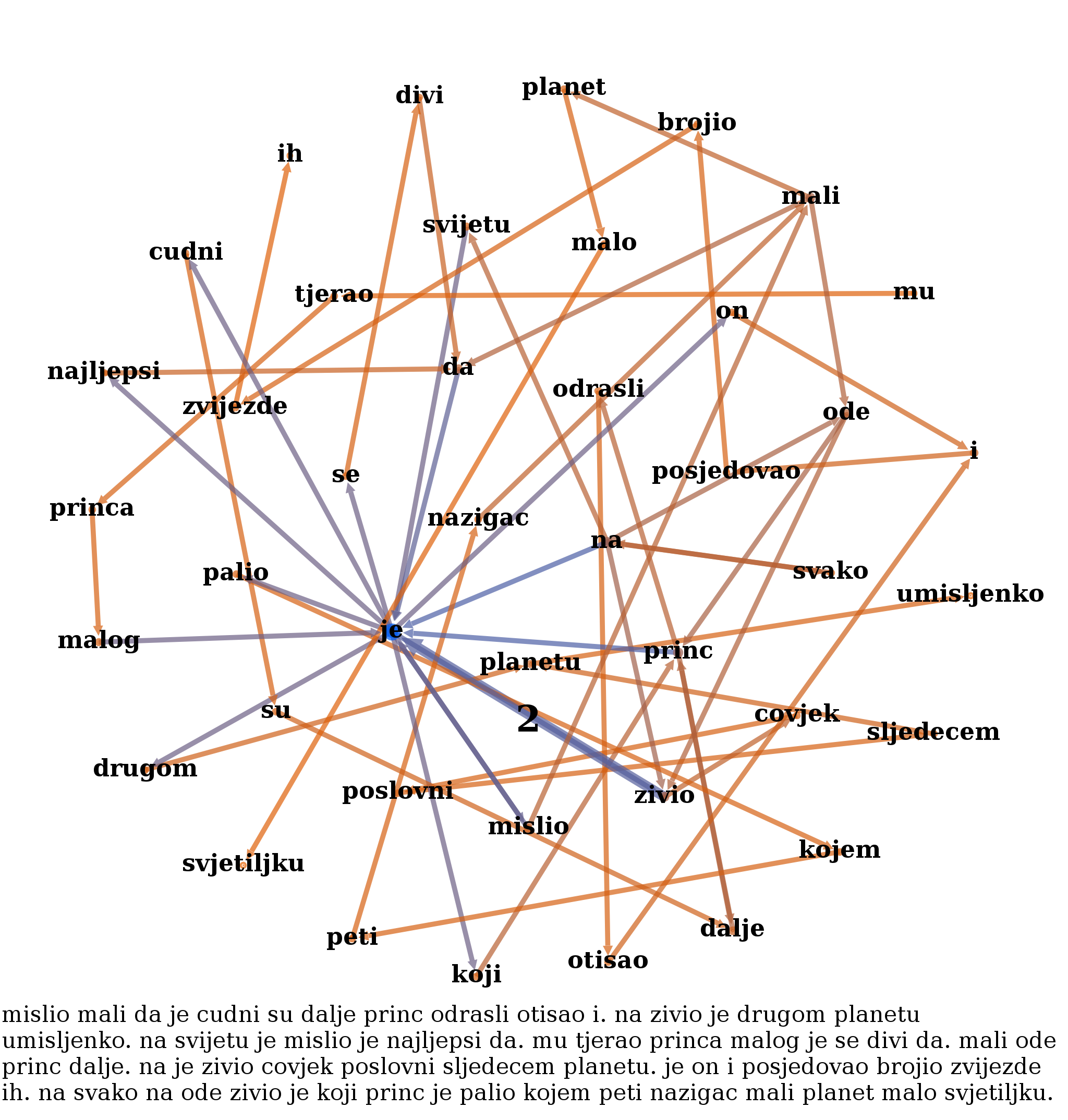}
\caption{Example network constructed from sentence-level shuffled text}
\label{fig_sim}
\end{figure}

\begin{figure}[!t]
\centering
\includegraphics[width=3.5in]{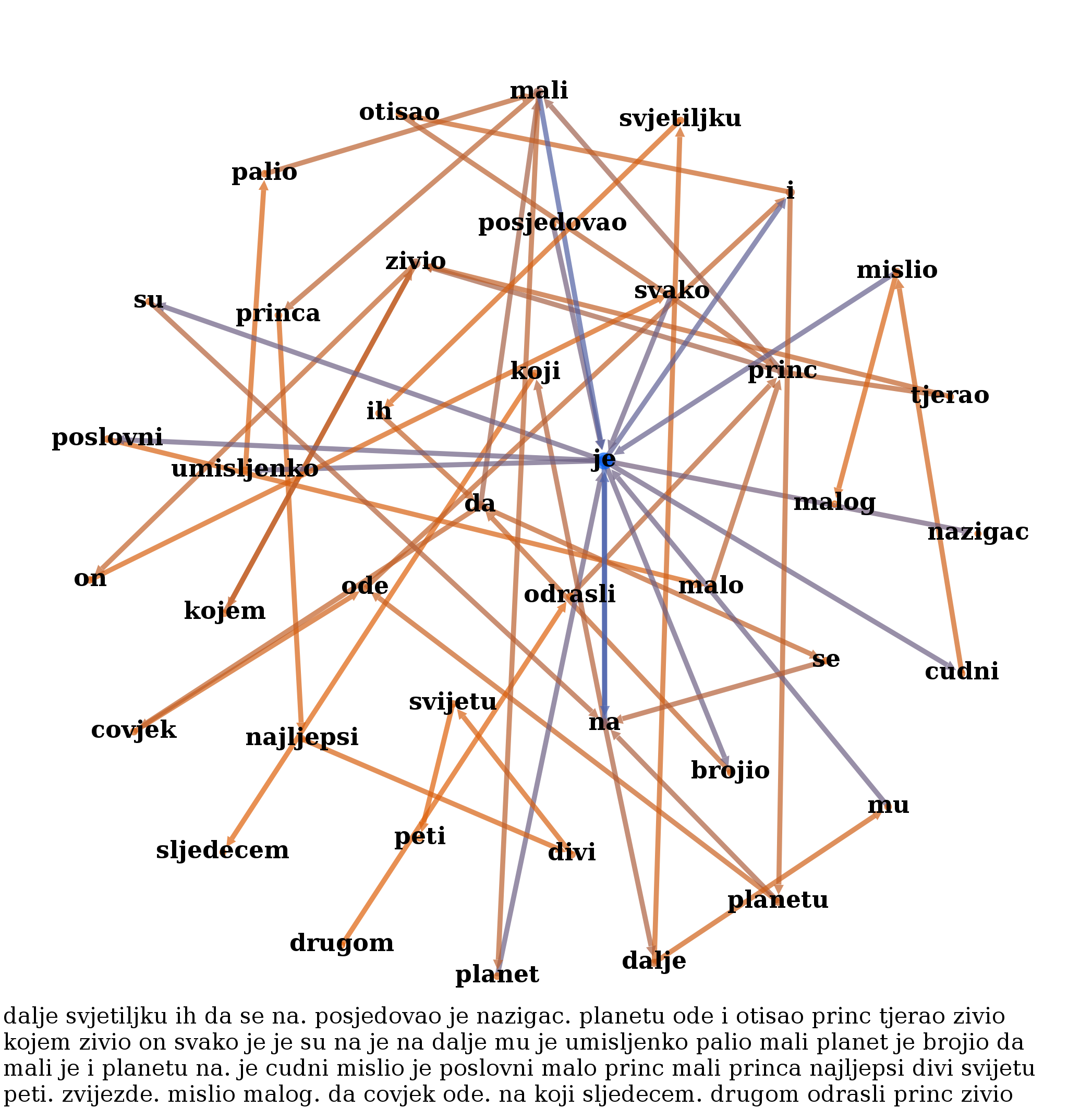}
\caption{Example network constructed from text-level shuffled text}
\label{fig_sim}
\end{figure}

\section{Results}
The measures of the networks constructed from C2 corpus, sentence-level shuffled C2', and text-level shuffled C2* are presented in Table 1. The results show that the shuffled networks have decreased values of the average path length $L$ and network diameter $D$, and have the value of the average clustering coefficient $C$ increased.

In the networks constructed from the corpora shuffled on the text-level the number of nodes $N$ ($N_{C1*} < N_{C1}, N_{C2*} < N_{C2}$) is smaller than the number of words in the original corpora. During the network construction process, the sentences containing just one word are disregarded, because our approach limits the word linkage to the sentence delimiters. This causes sentences with exactly one word to be isolated from the network, which reduces the number of nodes $N$ in C1* and C2*. 

\begin{table}
\caption{Network measures for original C2, sentence-level shuffled C2' and text-level shuffled C2*}
\begin{center}
\begin{tabular}{|l||c||c||c|}
\hline & C2 & C2' & C2* \\ \hline 
$N$ & 91328 & 91328 & 91204 \\ 
$K$ & 465196 & 586110 & 599335 \\ 
$L$ & 3.097 & 3.037 & 2.997 \\ 	
$D$ & 23 & 17 & 10 \\ 
$C$ & 0.317 & 0.343 & 0.354 \\ 
$\omega$ & 22 & 22 & 10 \\
\hline
\end{tabular}
\end{center}
\end{table}

One of the standard approaches to examine properties of co-occurrence networks is node degree distribution for unweighted and node strength distribution for weighted. Shuffling preserves the degree distribution due to the word frequencies  \cite{masucci2009differences} \cite{margan2014shuffled}. In this work we use rank distribution to compare different texts. Rank and frequency distributions are related through Zipf's law: the word frequency is inversely proportional to it's rank.

\begin{figure}[!t]
\centering
\includegraphics[width=3.5in]{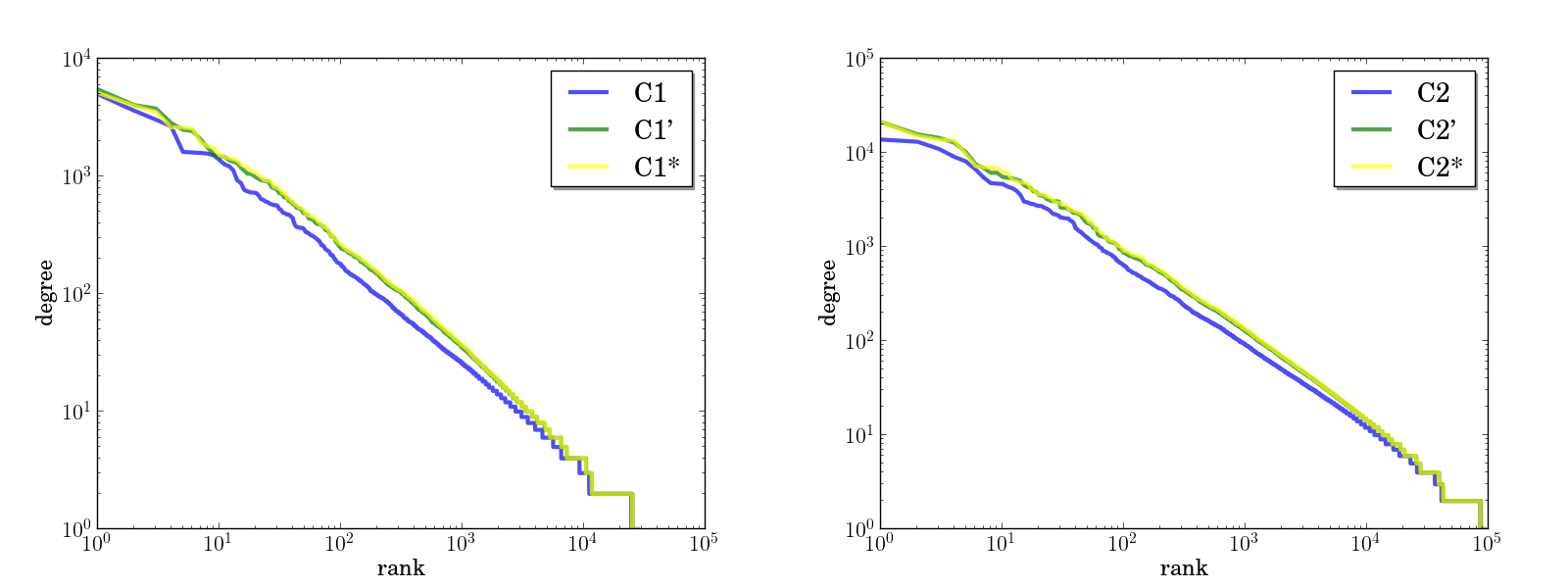}
\caption{Degree rank distributions for the C1, C1', C1*, C2, C2', C2*}
\label{fig_sim}
\end{figure}

\begin{figure}[!t]
\centering
\includegraphics[width=3.5in]{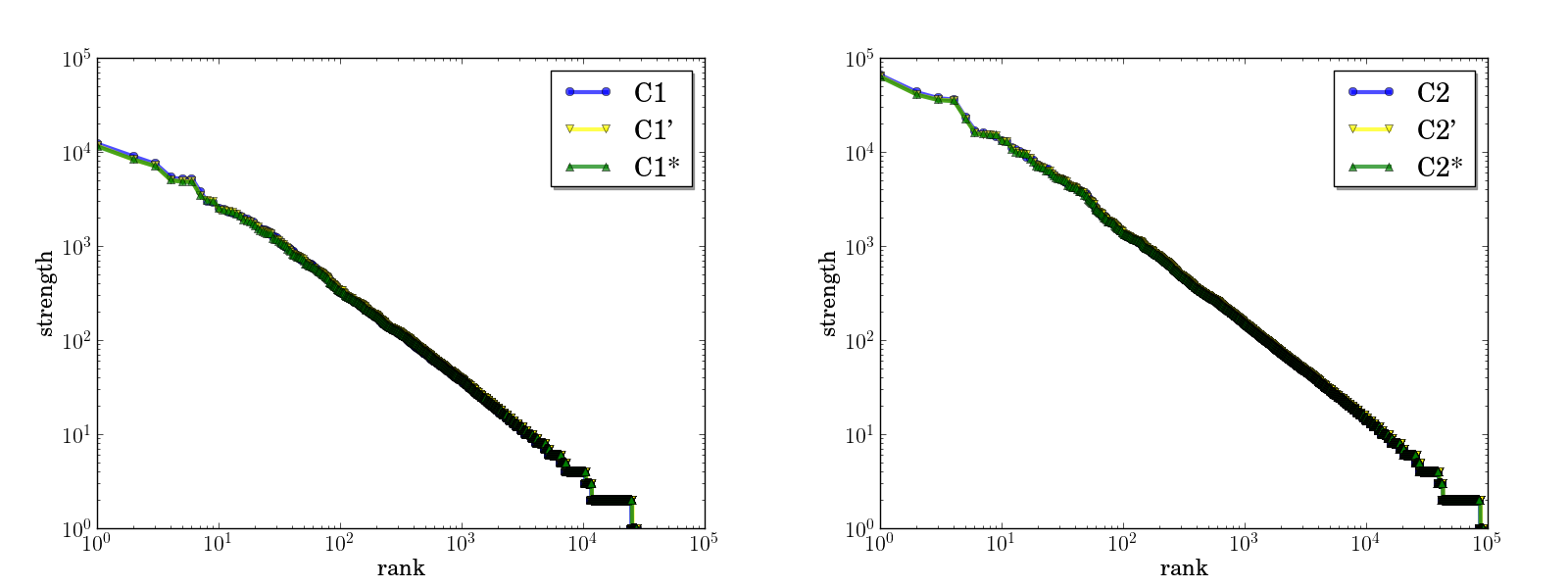}
\caption{Strength rank distributions for the C1, C1', C1*, C2, C2', C2*}
\label{fig_sim}
\end{figure}

Initially, we computed the degree and strength rank distributions only for C2 and corresponding shuffled versions. As expected, the degree rank distributions, as well as the strength rank distributions are preserved during the shuffling procedure. In order to explore whether the same holds for substantially smaller corpus we checked degree and strength rank distributions for C1 as well. Fig. 5 and Fig. 6 present degree and strength rank distributions for the C1, C1', C1*, C2, C2' and C2*.  Degree rank distributions exhibit no substantial deviation in shuffled networks, and strength distribution is preserved due to the same word frequencies. At this point all network measures and distributions showed no substantial differences between the structure of the original and meaningless texts. 

The global network measures: average shortest path length, diameter, clustering coefficient and degree and strength distribution may not be well-suited to discriminate between original and meaningless texts. Therefore, it is necessary to include local (node level) network measures. Motivated by the reported results form \cite{masucci2009differences}, which introduced the local measure of node selectivity, we applied the same principle on our data. In weighted and directed co-occurrence networks nodes have ingoing and outgoing links, therefore it is necessary to consider the in- and out- selectivity.

Fig. 7 and Fig. 8 present obtained  in- and out- selectivity ranks for C1 and C2, and their shuffled counterparts. Regardless of the links direction (in or out), selectivity values of networks from the shuffled texts are constantly below selectivity values calculated from the normal texts. This can be explained by the well known properties found in text - collocations. Collocations are typical word pairs, triples,...etc. which are recognized as standard phrases or names (e.g. New York). Shuffling procedures rearranged the word order, which  disassembled  collocations. This effect is reflected in the selectivity, since the high weights of common phrases and collocations are lost, which flattens and abates the selectivity ranks. 

\begin{figure}[!t]
\centering
\includegraphics[width=3.5in]{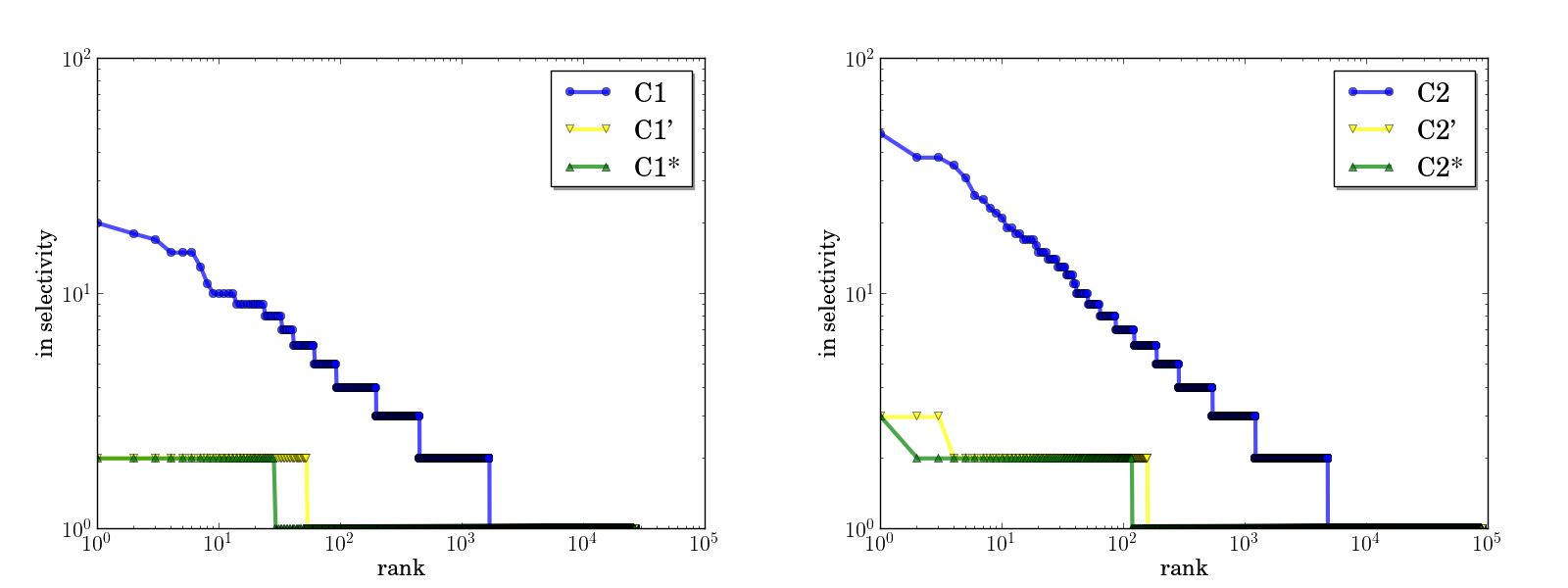}
\caption{In-selectivity rank distributions for the C1, C1', C1*, C2, C2', C2*}
\label{fig_sim}
\end{figure}

\begin{figure}[!t]
\centering
\includegraphics[width=3.5in]{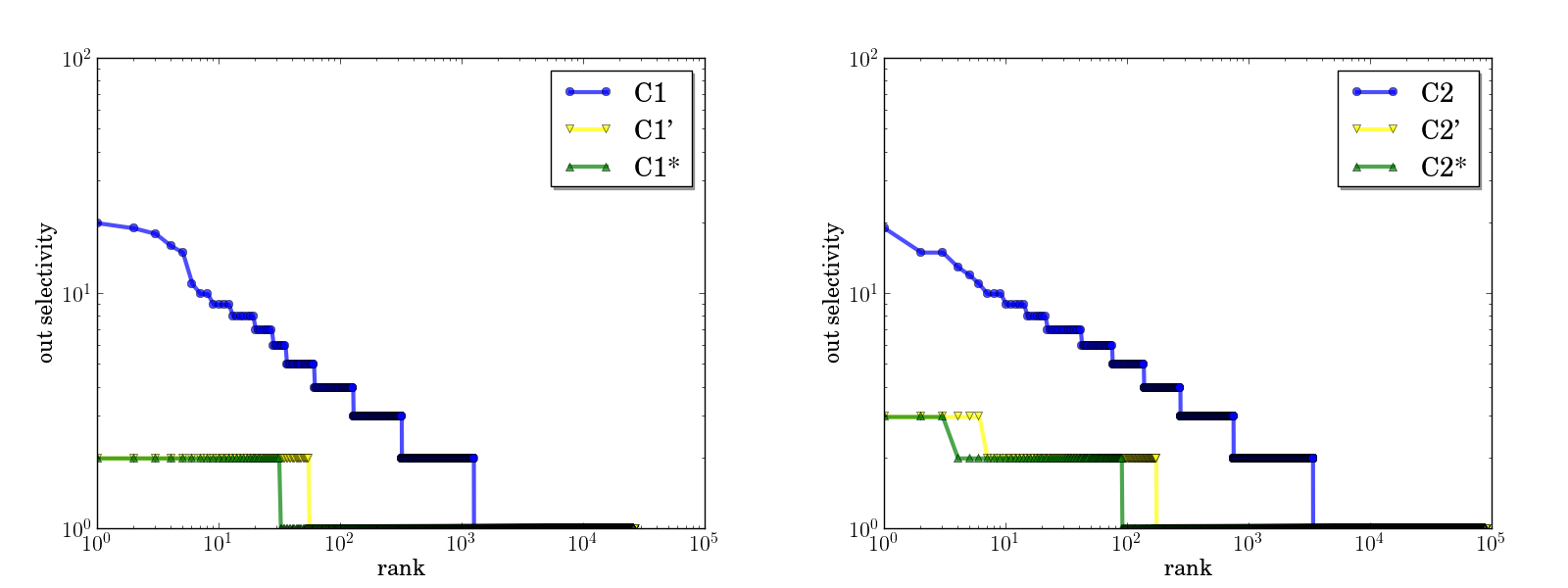}
\caption{Out-selectivity rank distributions for the C1, C1', C1*, C2, C2', C2*}
\label{fig_sim}
\end{figure}

\section{Conclusion}
We studied the structure of the linguistic networks constructed from normal and shuffled texts in Croatian.
As expected, the text shuffling causes the decrease of the average path length $L$ and the network diameter $D$, and the increase of the average clustering coefficient $C$.

We showed that the degree rank distributions exhibit no substantial deviation in shuffled networks, and the strength rank distributions are preserved due to the same word frequencies. The standard approach to study the structure of linguistic co-occurrence networks showed no substantial differences among the topologies of the original and shuffled texts. Therefore we utilized the node selectivity measure proposed by Masucci and Rodgers in \cite{masucci2009differences}.

Our results showed that the in- and out- selectivity values from shuffled texts are constantly below selectivity values calculated from normal texts. It seems that selectivity captures typical word phrases and collocations in Croatian which are lost during the shuffling procedure. The same holds for English where Masucci and Rodgers found that selectivity somehow captures the specialized local structures in nodes' neighborhood and forms of the morphological structures in text. Based on this findings, the measure of selectivity can be useful to discriminate between different text types, which will be the part of our future work. 

We have shown that the Croatian language networks have similar properties as language networks from English and other languages. Firstly, Croatian text shuffling has no influence on the degree and strength distributions, which has already been shown for English \cite{masucci2006network,masucci2009differences}, English and Portuguese \cite{caldeira2006network} and English, French, Spanish and Chinese \cite{krishna2011effect}. Furthermore, distance measures (average shortest path length and diameter) show that networks based on normal texts have a greater $L$ and $D$ value than the corresponding network based on shuffled text. The same relations for average clustering coefficient, average shortest path length and diameter are shown in \cite{krishna2011effect} for all studied languages (English, French, Spanish and Chinese). Similar results are shown for English and Portuguese in \cite{caldeira2006network}, although the authors used different shuffling procedures. 

Our results imply that the node selectivity is a measure suitable for fine-grained differentiation between an original and meaningless text. Furthermore, selectivity can potentially be considered as a measure for discrimination between different text types, capturing the aspect of quality of texts. This should be thoroughly examined in the future work, which will cover: a) differentiation between text types from linguistic network structure, b) finding which set of measures reflects the quality of the texts, c) the analysis of the Croatian linguistic networks using syntax dependencies instead of pure co-occurrence.

\end{document}